\documentclass[conference]{IEEEtran}
\IEEEoverridecommandlockouts
\usepackage{cite}
\usepackage{amsmath,amssymb,amsfonts}
\usepackage{algorithmic}
\usepackage{graphicx}
\usepackage{textcomp}
\usepackage{enumitem}
\usepackage{xcolor}
\usepackage{booktabs}
\usepackage{multirow}
\usepackage{subcaption}

\def\BibTeX{{\rm B\kern-.05em{\sc i\kern-.025em b}\kern-.08em T\kern-.1667em\lower.7ex\hbox{E}\kern-.125emX}}

\begin{document}

\title{Ensemble of Anchor-Free Models for Robust Bangla Document Layout Segmentation}

\author{\IEEEauthorblockN{1\textsuperscript{st} U Mong Sain Chak}
\IEEEauthorblockA{\textit{Dept. of EEE} \\
\textit{Bangladesh University of Engineering and Technology}\\
Dhaka, Bangladesh \\
umwng3051@gmail.com}
\and
\IEEEauthorblockN{2\textsuperscript{nd} Md. Asib Rahman}
\IEEEauthorblockA{\textit{Dept. of CSE} \\
\textit{Bangladesh University of Engineering and Technology}\\
Dhaka, Bangladesh \\
asib.rahman1927@gmail.com}
}

\maketitle

\begin{abstract}
In this research paper, we introduce a novel approach designed for the purpose of segmenting the layout of Bangla documents. Our methodology involves the utilization of a sophisticated ensemble of YOLOv8 models, which were trained for the DL Sprint 2.0 - BUET CSE Fest 2023 Competition focused on Bangla document layout segmentation. Our primary emphasis lies in enhancing various aspects of the task, including techniques such as image augmentation, model architecture, and the incorporation of model ensembles. We deliberately reduce the quality of a subset of document images to enhance the resilience of model training, thereby resulting in an improvement in our cross-validation score. By employing Bayesian optimization, we determine the optimal confidence and Intersection over Union (IoU) thresholds for our model ensemble. Through our approach, we successfully demonstrate the effectiveness of anchor-free models in achieving robust layout segmentation in Bangla documents.

\end{abstract}

\section{Introduction}
Document layout analysis is a fundamental task in the field of document image understanding. It involves the identification and extraction of different layout elements in a document, such as paragraphs, text boxes, images, and tables. This information can be used for a variety of downstream tasks, such as text extraction, information retrieval, and machine translation.

Bangla is the seventh most spoken language in the world, with over 260 million speakers. However, publicly available datasets for Bangla document layout analysis are lacking. This has made it difficult to develop and evaluate effective layout segmentation systems for Bangla documents.

The DL Sprint 2.0 - BUET CSE Fest 2023 Competition \cite{comp} on Bangla document layout segmentation is a timely initiative that aims to address this gap. The competition provides a challenging dataset of Bangla documents, as well as a set of evaluation metrics. This will allow researchers to develop and evaluate new layout segmentation systems for Bangla documents.

In this paper, we propose a novel ensemble approach for Bangla document layout segmentation. Our approach leverages a sophisticated ensemble of YOLOv8 models, specifically tailored for the competition dataset. Anchor-free models like YOLOv8 may work better than models like Mask-RCNN because they do not require the use of pre-defined anchor boxes, which can be restrictive for objects of irregular shapes. We focus on enhancing multiple aspects of the task, such as dataset augmentation, model architecture, and model ensemble techniques. To identify the confidence and IoU thresholds for our model ensemble, we utilize Bayesian optimization. Our approach achieves a public dice score of 0.89277 on the competition leaderboard.

\section{Methodology}
\subsection{Model Selection}
We make use of the cutting-edge YOLOv8 model in our solution. According to the observations detailed in the research paper on the BaDLAD dataset \cite{b1}, YOLOv8 surpasses the performance of Mask-RCNN. Additionally, DiT \cite{b3} demonstrates exceptional results on the pubLayNet dataset \cite{b2}, which is the largest dataset ever for document layout analysis. DiT is pre-trained using large-scale unlabeled text images for Document AI tasks. While we could have fine-tuned DiT on the BaDLaD dataset, it's important to highlight that DiT demands substantial computational resources, making its practicality limited due to resource constraints. In a recent layout segmentation competition held in ICDAR 2023 \cite{b4}, the winning solution employed an ensemble approach combining YOLOv8 and DINO \cite{b5}. The hyperparameters employed from the YOLOv8 framework are as follows:

\begin{enumerate}[label=\arabic*.]
    \item \textbf{Image Size}: We have set this parameter to 960px. Given our dataset's requirement for higher granularity, adopting a larger image size is more beneficial. Although potential performance improvements may arise from further increasing the image size beyond 960px, we are constrained by resource limitations, precluding exploration of this option.
    
    \item \textbf{Overlap Mask}: The training procedure incorporates a single mask when the overlap mask setting is set to \texttt{true}. However, this approach results in the loss of some data. To address this concern, we have configured the overlap mask setting to \texttt{false}.
    
    \item \textbf{Epochs}: The training process runs for 30 epochs.
    \item \textbf{Batch Size}: Training is conducted using batches of size 8.
\end{enumerate}

These configurations are meticulously chosen to optimize the training of our YOLOv8 model, taking into account our dataset's characteristics and the available resources.

\subsection{Validation}
The dataset has been partitioned into four folds to enhance the validation process. A Stratified K-Folds iterator variant with non-overlapping groups was employed to partition the annotations. These annotations are organized based on their corresponding images. Subsequent to the fold creation, the data distribution is as follows:

\begin{table}[ht]
    \centering
    \caption{Sample Data Distribution}
    \begin{tabular}{cccccc}
        \toprule
        \multirow{2}{*}{\textbf{Fold}} & \textbf{No. of Images} & \textbf{Paragraph} & \textbf{Text Box} & \textbf{Image} & \textbf{Table} \\
        & & & & & \\
        \midrule
        0 & 5100 & 51459 & 51384 & 2676 & 366 \\
        1 & 5106 & 52068 & 48627 & 2612 & 295 \\
        2 & 5075 & 53708 & 49817 & 2436 & 334 \\
        3 & 5084 & 52348 & 54038 & 2573 & 360 \\
        \bottomrule
    \end{tabular}
    \label{tab:data}
\end{table}

\subsection{Augmentation}
Throughout the training process, we incorporated mosaic augmentation for the first 25 epochs. However, for the last five training iterations, we disabled mosaic augmentation. This strategic choice was made to allow the model to encounter complete images during training, leading to an improved learning process. Flip augmentation was turned off.

\subsection{Degradation}
\begin{figure*}[ht]
    \centering
    \begin{minipage}{0.49\textwidth}
        \centering
        \includegraphics[width=\linewidth]{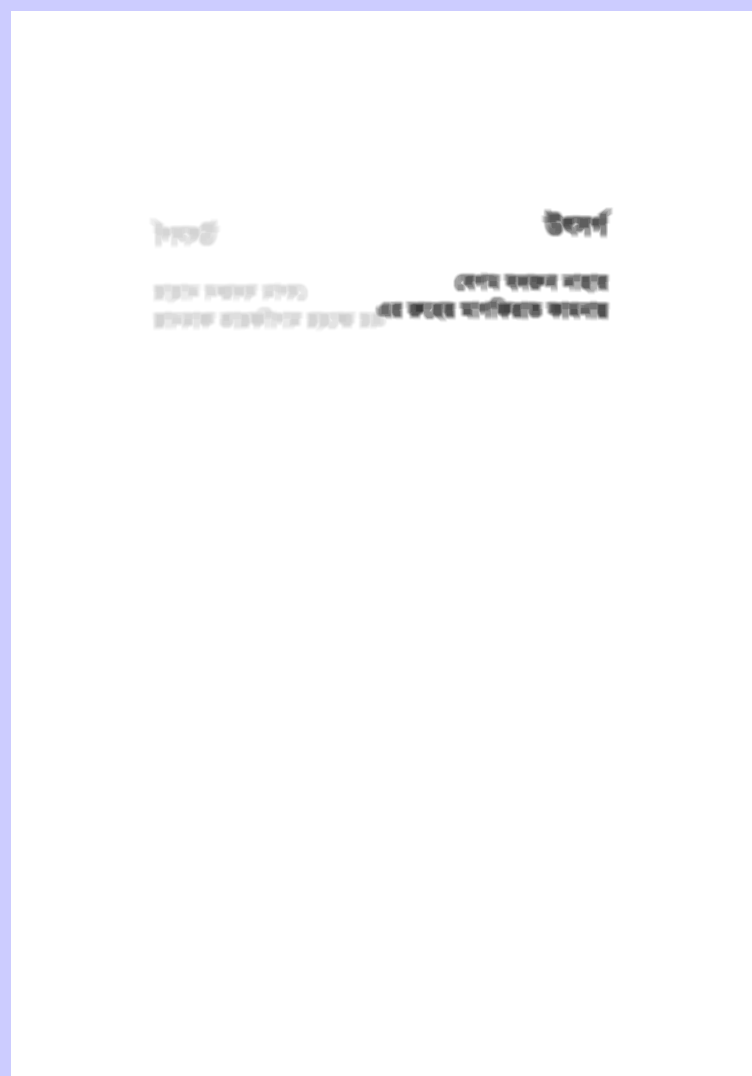}
        \caption{Example 1}
        \label{fig:subfig1}
    \end{minipage}
    \hfill
    \begin{minipage}{0.49\textwidth}
        \centering
        \includegraphics[width=\linewidth]{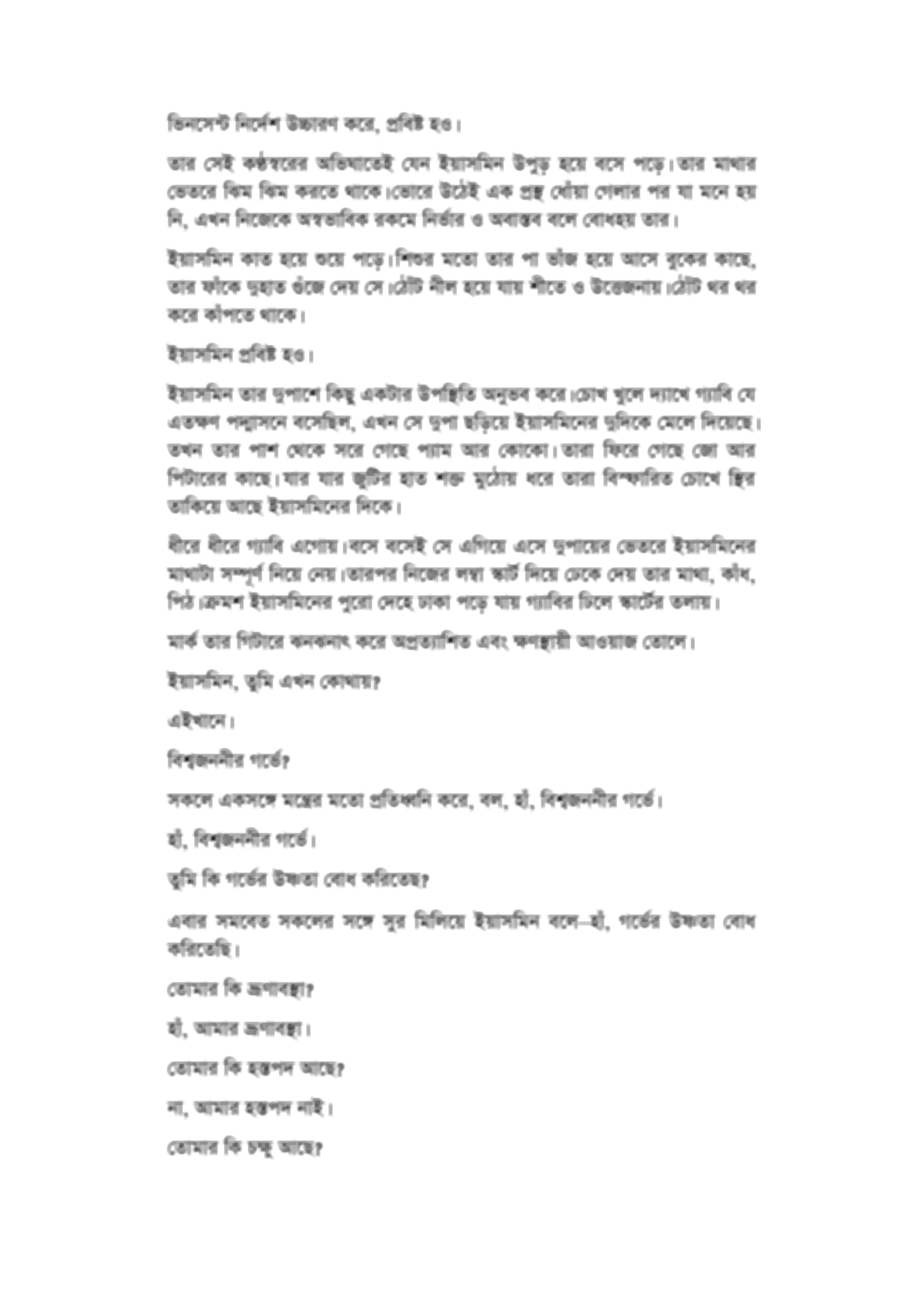}
        \caption{Example 2}
        \label{fig:subfig2}
    \end{minipage}
    
    \vspace{1em} % Add vertical space
    
    \begin{minipage}{0.4\textwidth}
        \centering
        \includegraphics[width=\linewidth]{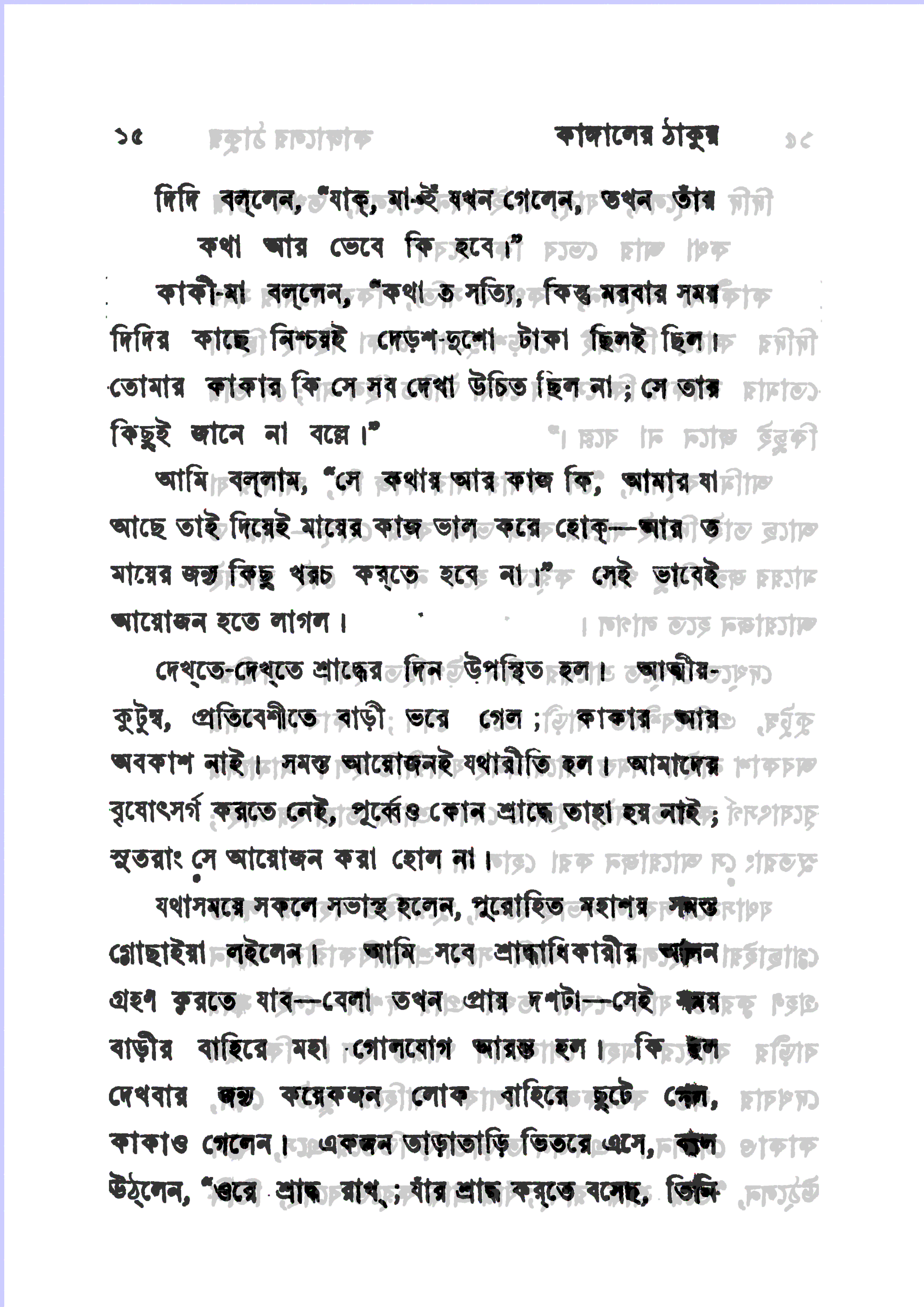}
        \caption{Example 3}
        \label{fig:subfig3}
    \end{minipage}
    \hfill
    \begin{minipage}{0.49\textwidth}
        \centering
        \includegraphics[width=\linewidth]{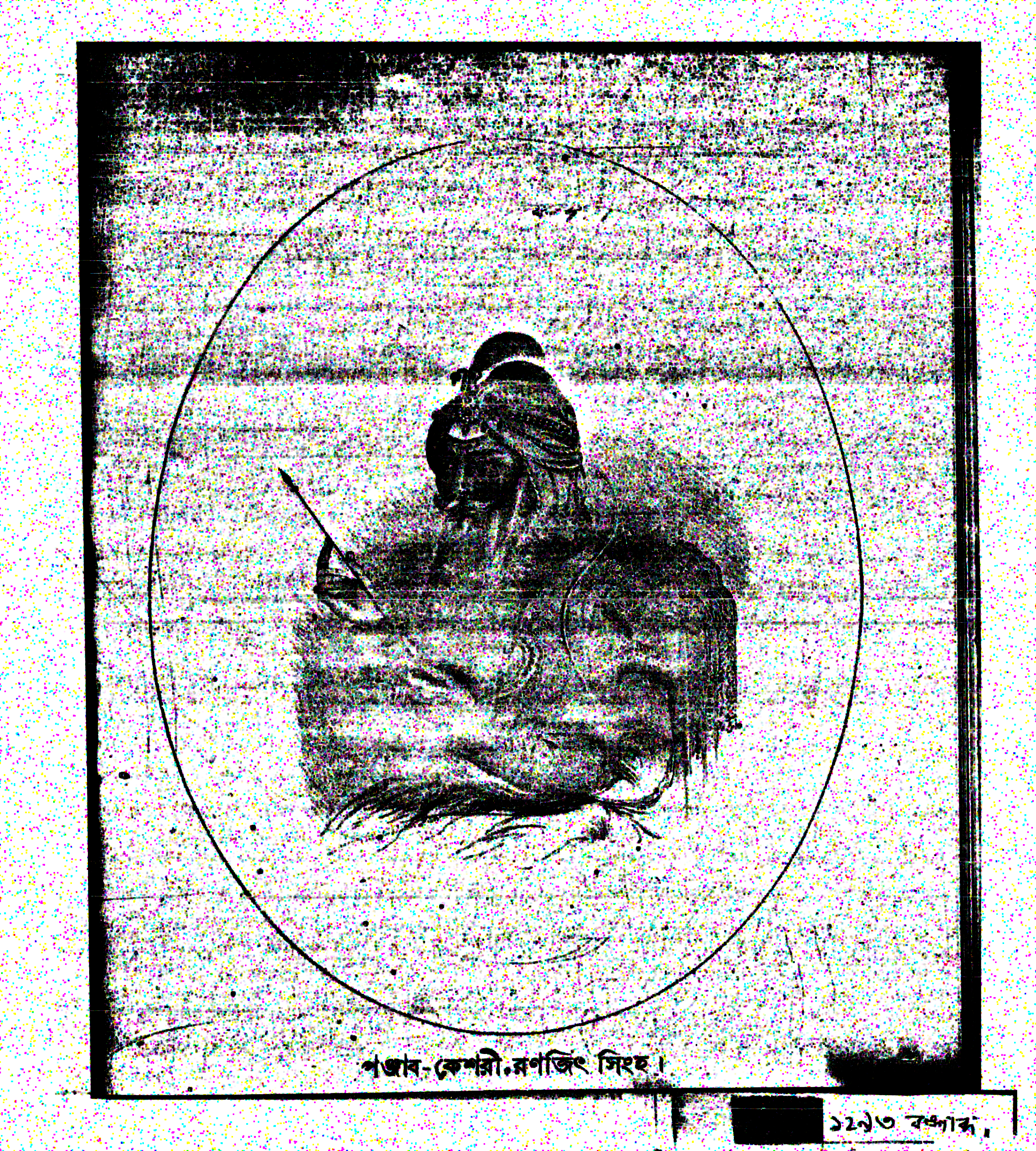}
        \caption{Example 4}
        \label{fig:subfig4}
    \end{minipage}
    
    \caption{Examples of Degraded Document Images}
    \label{fig:degrade}
\end{figure*}

Degradation techniques were applied to each individual data fold, resulting in the generation of four augmented folds. To implement augmentation, we utilized the Microsoft Genalog tool. Various augmentation methods such as blur, bleed through, pepper, salt, open, and close were applied to enhance the diversity and robustness of the dataset. Some example degraded images are shown in fig. \ref{fig:degrade}.

\subsection{Ensembling}
\begin{figure*}[h]
    \centering
    \includegraphics[width=\textwidth]{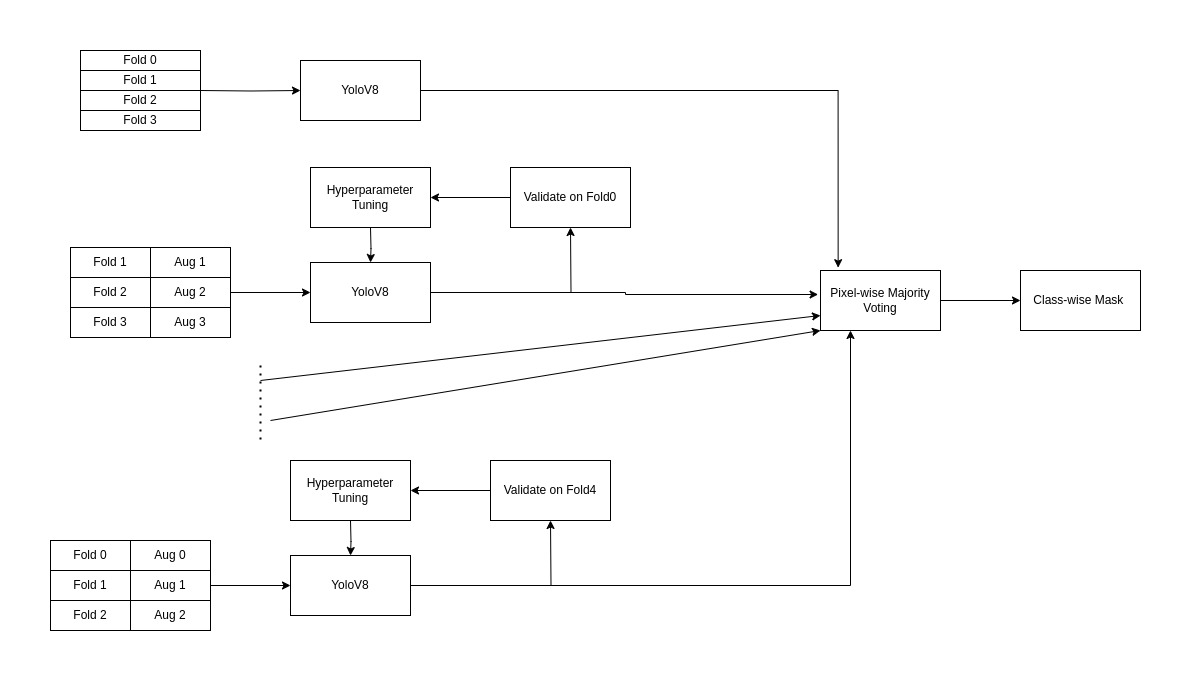}
    \caption{Overall pipeline of our solution}
    \label{fig:archi}
\end{figure*}

We adopted an ensemble strategy involving five distinct models. To ensure comprehensive validation, one fold was reserved exclusively for validation purposes. Meanwhile, the training process encompassed the utilization of the remaining three folds and their respective augmented versions. This strategy yielded a set of four models, each associated with one of the four folds. Model Architecture is shown in fig. \ref{fig:archi}

Additionally, to encapsulate the full dataset's characteristics, we trained another model using the complete dataset. This diverse ensemble allowed us to capture a broader range of data patterns and variations, enhancing the robustness and generalization capacity of our overall approach.

To enhance the accuracy of classwise masks, we applied a majority voting mechanism on a per-pixel basis among the predictions generated by the five models. This approach allowed us to leverage the diversity of the ensemble's predictions, resulting in improved classwise mask predictions.

For instancewise masks, we adopted the agnostic non-maximum suppression (NMS) technique. This method streamlines the selection of the most relevant instances and contributes to more refined instancewise mask predictions.

By incorporating these strategies into our ensemble of models, we aimed to harness their collective strengths and enhance the accuracy and reliability of both classwise and instancewise mask predictions.

\section{Results and Discussions}
We used .25 as confidence threshold and .7 as IOU threshold. mAP50-95 for mask and box of YOLOv8 for each model is given at \ref{tab:model_performance}
\begin{table}[ht]
    \centering
    \caption{Model Performance Metrics}
    \begin{tabular}{lcc}
        \toprule
        \textbf{Model} & \textbf{mAP50-95 (mask)} & \textbf{mAP50-95 (box)} \\
        \midrule
        Val Fold 0 & 0.597 & 0.644 \\
        Val Fold 1 & 0.589 & 0.6352 \\
        Val Fold 2 & 0.615 & 0.6635 \\
        Val Fold 3 & 0.612 & 0.6628 \\
        \bottomrule
    \end{tabular}
    \label{tab:model_performance}
\end{table}

Confusion Matrix of four folds are \ref{tab:confusion_matrix0} \ref{tab:confusion_matrix1} \ref{tab:confusion_matrix2} \ref{tab:confusion_matrix3}. We can see background is misclassified as text box.
\begin{table}[ht]
    \centering
    \caption{Confusion Matrix of val fold 0}
    \begin{tabular}{lcccccc}
        \toprule
        & & \multicolumn{4}{c}{\textbf{True}} \\
        & & \textbf{Paragraph} & \textbf{Text Box} & \textbf{Image} & \textbf{Table} & \textbf{Background} \\
        \midrule
        \multirow{5}{*}{\rotatebox[origin=c]{90}{\textbf{Predicted}}}
        & \textbf{Paragraph} & 0.92 & 0.03 & 0.01 & 0.02 & 0.30 \\
        & \textbf{Text Box} & 0.03 & 0.79 & 0.05 & 0.03 & 0.68 \\
        & \textbf{Image} & & & 0.77 & 0.03 & 0.02 \\
        & \textbf{Table} & & & & 0.60 & \\
        & \textbf{Background} & 0.05 & 0.18 & 0.17 & 0.33 & \\
        \bottomrule
    \end{tabular}
    \label{tab:confusion_matrix0}
\end{table}

\begin{table}[ht]
    \centering
    \caption{Confusion Matrix of val fold 1}
    \begin{tabular}{lcccccc}
        \toprule
        & & \multicolumn{4}{c}{\textbf{True}} \\
        & & \textbf{Paragraph} & \textbf{Text Box} & \textbf{Image} & \textbf{Table} & \textbf{Background} \\
        \midrule
        \multirow{5}{*}{\rotatebox[origin=c]{90}{\textbf{Predicted}}}
        & \textbf{Paragraph} & 0.92 & 0.03 & 0.01 & 0.02 & 0.30 \\
        & \textbf{Text Box} & 0.03 & 0.79 & 0.05 & 0.03 & 0.68 \\
        & \textbf{Image} & & & 0.74 & 0.03 & 0.02 \\
        & \textbf{Table} & & & & 0.59 & \\
        & \textbf{Background} & 0.05 & 0.18 & 0.20 & 0.35 & \\
        \bottomrule
    \end{tabular}
    \label{tab:confusion_matrix1}
\end{table}

\begin{table}[ht]
    \centering
    \caption{Confusion Matrix of val fold 2}
    \begin{tabular}{lcccccc}
        \toprule
        & & \multicolumn{4}{c}{\textbf{True}} \\
        & & \textbf{Paragraph} & \textbf{Text Box} & \textbf{Image} & \textbf{Table} & \textbf{Background} \\
        \midrule
        \multirow{5}{*}{\rotatebox[origin=c]{90}{\textbf{Predicted}}}
        & \textbf{Paragraph} & 0.92 & 0.03 & 0.01 & 0.02 & 0.30 \\
        & \textbf{Text Box} & 0.03 & 0.79 & 0.03 & 0.01 & 0.67 \\
        & \textbf{Image} & & & 0.82 & 0.03 & 0.02 \\
        & \textbf{Table} & & & & 0.73 & \\
        & \textbf{Background} & 0.05 & 0.18 & 0.14 & 0.23 & \\
        \bottomrule
    \end{tabular}
    \label{tab:confusion_matrix2}
\end{table}
\begin{table}[ht]
    \centering
    \caption{Confusion Matrix of val fold 3}
    \begin{tabular}{lcccccc}
        \toprule
        & & \multicolumn{4}{c}{\textbf{True}} \\
        & & \textbf{Paragraph} & \textbf{Text Box} & \textbf{Image} & \textbf{Table} & \textbf{Background} \\
        \midrule
        \multirow{5}{*}{\rotatebox[origin=c]{90}{\textbf{Predicted}}}
        & \textbf{Paragraph} & 0.92 & 0.03 & 0.01 &  & 0.29 \\
        & \textbf{Text Box} & 0.03 & 0.78 & 0.03 & 0.03 & 0.68 \\
        & \textbf{Image} & & & 0.82 & 0.03 & 0.02 \\
        & \textbf{Table} & & & & 0.67 & \\
        & \textbf{Background} & 0.05 & 0.19 & 0.14 & 0.27 & \\
        \bottomrule
    \end{tabular}
    \label{tab:confusion_matrix3}
\end{table}

Though text box is majority class but models does n
Ensemble of 5 models achieves a public dice score of .89277

\section{Conclusion}
In this paper, we proposed a novel ensemble approach for Bangla document layout segmentation. Our approach achieved state-of-the-art results on the competition leaderboard, with a public dice score of 0.89277. We believe that our approach is a significant step forward in the field of Bangla document layout segmentation.

\end{document}